\documentclass[Afour,sageh,times,doublespace]{sagej}

\usepackage{moreverb,url}

\usepackage[colorlinks,bookmarksopen,bookmarksnumbered,citecolor=red,urlcolor=red]{hyperref}

\newcommand\BibTeX{{\rmfamily B\kern-.05em \textsc{i\kern-.025em b}\kern-.08em
T\kern-.1667em\lower.7ex\hbox{E}\kern-.125emX}}

\def\volumeyear{2016}

\usepackage[T1]{fontenc}
\usepackage{times}
\usepackage{soul}
\usepackage{url}
\usepackage{booktabs}
\usepackage{algorithm}
\usepackage{algorithmic}
\usepackage{multirow}
\usepackage{array}
\usepackage{xcolor}
\usepackage{float}

\definecolor{purple}{RGB}{153, 0, 255}
\DeclareMathOperator*{\argmax}{arg\,max}

\usepackage{hhline}
\usepackage[caption=false]{subfig}
\usepackage{listings}
\usepackage{fancyvrb}
\usepackage{xcolor}
\definecolor{highlight}{HTML}{DAE8FC}
\definecolor{blue}{HTML}{4a86e8}
\definecolor{yellow}{HTML}{e69138}
\definecolor{pink}{HTML}{e06666}
\newcommand{\method}[0]{\textsc{LLM-GROP}}

\newcommand{\random}{TPRA}
\newcommand{\LLM}{LATP}
\usepackage{threeparttable}
\usepackage{caption}
\begin{document}

\runninghead{Authors' last name}

\title{\method{}: Visually Grounded Robot Task and Motion Planning with Large Language Models}

\author{Xiaohan Zhang*\affilnum{1},  
Yan Ding*\affilnum{1,5,6}, 
Yohei Hayamizu*\affilnum{1}, 
Zainab Altaweel*\affilnum{1}, 
Yifeng Zhu\affilnum{2}, \\
Yuke Zhu\affilnum{2}, 
Peter Stone\affilnum{2,3}, 
Chris Paxton\affilnum{4}, 
Shiqi Zhang\affilnum{1}
}

\affiliation{\affilnum{1}The State University of New York at Binghamton\\
\affilnum{2}The University of Texas at Austin\\
\affilnum{3}Sony AI\\
\affilnum{4}Hello Robot\\
\affilnum{5}Shanghai AI Laboratory\\
\affilnum{6}OneStar Robotics\\
\affilnum{*}Equal contribution\\
}

\corrauth{Shiqi Zhang}

\email{zhangs@binghamton.edu}

\begin{abstract}
Task planning and motion planning are two of the most important problems in robotics, where task planning methods help robots achieve high-level goals and motion planning methods maintain low-level feasibility. 
Task and motion planning (TAMP) methods interleave the two processes of task planning and motion planning to ensure goal achievement and motion feasibility. 
Within the TAMP context, we are concerned with the mobile manipulation (MoMa) of multiple objects, where it is necessary to interleave actions for navigation and manipulation. 

In particular, we aim to compute where and how each object should be placed given underspecified goals, such as ``set up dinner table with a fork, knife and plate.''
We leverage the rich common sense knowledge from large language models (LLMs), e.g., about how tableware is organized, to facilitate both task-level and motion-level planning. 
In addition, we use computer vision methods to learn a strategy for selecting base positions to facilitate MoMa behaviors, where the base position corresponds to the robot's ``footprint'' and orientation in its operating space. 
Altogether, this article provides a principled TAMP framework for MoMa tasks that accounts for common sense about object rearrangement and is adaptive to novel situations that include many objects that need to be moved. 
We performed quantitative experiments in both real-world settings and simulated environments. 
We evaluated the success rate and efficiency in completing long-horizon object rearrangement tasks. 
While the robot completed 84.4\% real-world object rearrangement trials, subjective human evaluations indicated that the robot's performance is still lower than experienced human waiters. 
\end{abstract}


\maketitle

\section{Introduction}

Robots require task planning methods to sequence symbolic actions for accomplishing complex tasks.
They also need motion planning methods to compute trajectories that realize these symbolic actions while ensuring motion-level feasibility.
Task and motion planning (TAMP) refers to a family of algorithms that integrate task and motion planning processes to compute motion trajectories that can be directly executed on robot hardware to achieve task-level goals~\citep{garrett2021integrated,zhao2024survey}.
While most existing TAMP algorithms are designed for purely manipulation (i.e., requiring no navigation) domains, robots may need to handle objects located far apart, requiring a combination of navigation and manipulation actions.
In this article, we study \emph{mobile manipulation} (MoMa) domains in which  robots perform both navigation and manipulation tasks.
This article focuses on addressing MoMa challenges by developing TAMP methods that are visually grounded and capable of leveraging common sense knowledge to achieve underspecified goals.

\begin{figure*}
  \begin{center}
    \includegraphics[width=\textwidth]{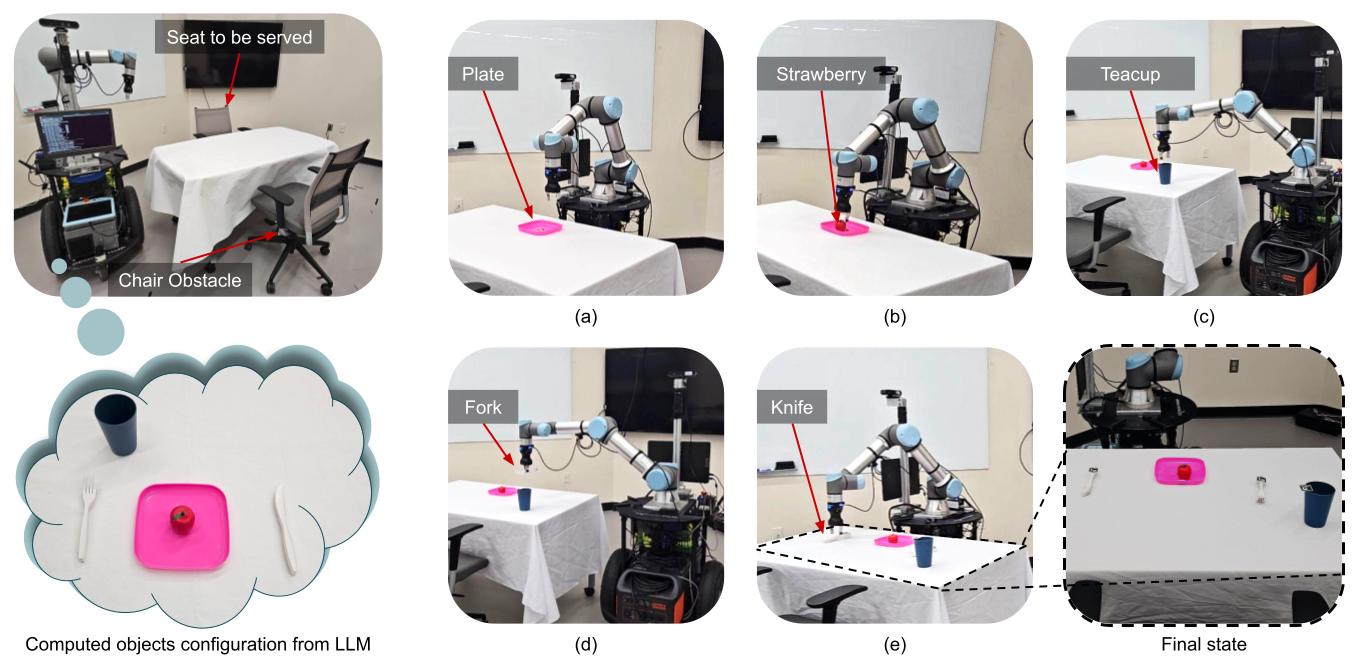}
    \caption{An illustration of our mobile manipulation (MoMa) domain, where a mobile manipulator is tasked with setting a dining table.
    The robot must arrange several tableware items, including a knife, a fork, a plate, a cup mat, and a mug.
    These objects are located on other tables, and the environment also includes randomly generated obstacles (e.g., chairs), which are not accounted for in the pre-built map.
    The robot must compute semantically specified goal configurations of the objects and (task and motion) plans for rearranging the objects on the target table. The computed plan includes both navigation and manipulation behaviors. 
    }
    \label{fig:set_table}
  \end{center}
\end{figure*}

Multi-object rearrangement is an essential skill for service robots to perform everyday tasks such as setting tables, organizing bookshelves, and loading dishwashers~\citep{habitatrearrangechallenge2022,RoomR}.
These tasks require robots to demonstrate both manipulation and navigation capabilities.
For instance, a robot tasked with setting a dinner table may need to retrieve tableware items like forks and knives from different locations and place them onto a table surrounded by chairs, as illustrated in Figure~\ref{fig:set_table}.
To complete this task, the robot must accurately position the tableware in semantically specified configurations (e.g., placing the fork to the left of the knife) and efficiently navigate indoor spaces while avoiding obstacles such as chairs or humans, whose locations are not known in advance.
A variety of mobile manipulation systems have been developed for object rearrangement tasks~\citep{goodwin2022semantically,liu2022structformer,wei2023lego,huang2019large,gu2022multi,king2016rearrangement,cheong2020relocate,vasilopoulos2021reactive}.
Most of these systems require explicit instructions, such as arranging similarly colored items in a line or placing them in a specific shape on a table~\citep{goodwin2022semantically,huang2019large,gu2022multi,cheong2020relocate,vasilopoulos2021reactive,liang2022code}.
However, real-world user requests are often underspecified; for example, there are many ways to set a table, but some are preferred more than others.
How does a robot determine that a fork should be placed to the left of a plate and a knife to the right?
Reasoning about such societal conventions requires substantial common sense knowledge. 
Existing research has shown that large language models (LLMs), such as ChatGPT~\citep{openai}, possess a significant amount of such common sense knowledge~\citep{liu2021pre}.
In the past, researchers have equipped mobile manipulators with semantic information using machine learning methods~\citep{liu2022structformer,liu2022structdiffusion,wei2023lego,zhang2021hierarchical}.
However, these methods rely on training data, limiting their applicability for robots performing diverse service tasks in open worlds, where data collection can be difficult. 

To compute semantically specified goal configurations and enable mobile manipulation for object rearrangement, we introduce \method{}, which stands for \textbf{L}arge \textbf{L}anguage \textbf{M}odel for \textbf{G}rounded \textbf{RO}bot Task and Motion \textbf{P}lanning, 
Our approach leverages common sense knowledge for planning object rearrangement tasks.
\method{} first uses an LLM to generate \emph{symbolic} spatial relationships among objects, e.g., placing a fork and a knife to the left and right of a plate, respectively.
These symbolic relationships are then mapped to \emph{geometric} spatial relationships, whose feasibility is evaluated by a motion planning system.
For example, some areas of a table may be more feasible for object placement than others.
Finally, computer vision methods are employed to optimize the feasibility and efficiency of task-motion plans, maximizing long-term utility by balancing motion feasibility and task-completion efficiency.

We applied \method{} in a dining domain where a mobile manipulator is tasked with setting a table based on user instructions.
The robot is provided with a set of tableware objects and must compute a tabletop configuration that adheres to common sense rules while also generating a task-motion plan to execute the arrangement.
To evaluate the performance of our approach, we collected user ratings of different table settings for subjective evaluation.
Our results showed that \method{} improved user satisfaction compared to existing object rearrangement methods while maintaining similar or lower cumulative action costs.
Additionally, \method{} was demonstrated and evaluated on a real robot.

This article builds on our previous research, which introduced the initial version of \method{}~\citep{ding2023task} and its vision component, GROP~\citep{zhang2022visually}.
GROP used visual perception to identify optimal standing positions, maximizing the feasibility and efficiency of both manipulation and navigation actions.
\method{} extends this by incorporating LLMs to generate tabletop configurations that adhere to common sense principles and are feasible for TAMP systems.
Compared to the two conference papers, the primary contribution of this article is the unification of these two algorithms and systems, providing a cohesive presentation of all \method{} components.
We have updated illustrative examples and overview figures for improved clarity (Figures~\ref{fig:set_table}, \ref{fig:overview}, and \ref{fig:trials_result}).
Additional experiments were conducted to evaluate \method{} using different LLMs (Table~\ref{table:ablation result}), on robot hardware (Figure~\ref{fig:sankey_graph}), and through subjective evaluations of real-robot performance with human participants (Table~\ref{table:real robot exp}).
Challenges and opportunities are discussed toward the end of the article.

\section{Related Work}
\label{sec:related}

In this section, we first summarize task and motion planning literature, then introduce the mobile manipulation problem in object rearrangement domains, and finally discuss foundation models for robot planning. 

\subsection{Task and Motion Planning (TAMP)}
TAMP methods aim to compute plans that fulfill task-level goals while maintaining motion-level feasibility, as reviewed in recent articles~\citep{lagriffoul2018platform,garrett2021integrated,zhao2024survey}. 
Several TAMP algorithms have been introduced in recent years (e.g.,~\citep{gravot2005asymov,plaku2007discrete,erdem2011combining,srivastava2014combined,lagriffoul2014efficiently, chitnis2016guided,garrett2018ffrob,wang2018active,kim2019learning,chitnis2019learning,ding2022learning,zhu2020hierarchical,kim2020learning, dantam2018incremental}.
We distinguish a few subareas of TAMP that are closest to our research on learning to visually ground symbolic spatial relationships towards planning efficient and feasible task-motion behaviors under uncertainty. 

When high-level actions only take a few seconds, TAMP algorithms can focus mostly on action feasibility constraints without fully optimizing high-level plan efficiency.
However, when there are actions that take significant time to execute (e.g., long-distance navigation), task-completion efficiency cannot be overlooked.
Some recent methods have considered efficiency in different aspects of TAMP, such as planning task-level optimal behaviors in navigation domains~\citep{lo2020petlon}, integrating reinforcement learning with symbolic planning in dynamic environments~\citep{jiang2019taskb}, computing safe and efficient plans for urban driving~\citep{ding2020task}, and optimizing robot navigation actions under the uncertainty from motion and sensing~\citep{thomas2021mptp}.
In contrast to those methods that do not have a perception component, the main difference is that \method{} visually grounds symbols (about spatial relationships) to probabilistically evaluate action feasibility for task-motion planning. 
Another difference is that \method{} leverage LLMs for computing semantically meaning goal configurations. 

While most TAMP methods assume a fully observable and deterministic world~\citep{garrett2021integrated}, some have been developed to account for the uncertainty from perception and action outcomes~\citep{kaelbling2013integrated,hadfield2015modular,phiquepal2019combined,garrett2020online, nouman2021hybrid, akbari2020contingent}. 
For instance, the work of Kaelbling and Lozano-P{\'e}rez extended the ``hierarchical planning in the now'' approach to address both current-state uncertainty and future-state uncertainty~\citep{kaelbling2013integrated}. 
Going beyond those methods that aim to maintain plan feasibility to complete tasks under high-level uncertainty, we consider uncertainty in the robot motion and also incorporate task-completion efficiency into the optimization of robot behaviors. 
As a result, our \method{} algorithm is particularly suitable for TAMP domains that require robot operations over extended periods of time, such as long-distance navigation. 

Existing research has shown that visual information can be used to help robots predict plan feasibility, including task-level feasibility~\citep{driess2020deepr,zhu2020hierarchical}, and motion-level feasibility~\citep{driess2020deeph,wells2019learning}. 
Those methods were developed to maximize task completion rate in manipulation domains, and actions that take relatively long time (such as long-distance navigation) were not included in their evaluations. 
\method{} incorporates efficiency into plan optimization, while leveraging common sense from LLMs for computing goal configurations. 
For instance, when highly feasible plans have very high costs, \method{} supports the flexibility of executing slightly less feasible plans with much lower costs. 
\method{} achieves this desirable trade-off between feasibility and efficiency by probabilistically evaluating plan feasibility, which is not supported by the above-mentioned methods. 

\subsection{MoMa for Object Rearrangement}
Rearranging objects is a critical task for service robots, and much research has focused on moving objects from one location to another and placing them in new positions. 
Examples include the Habitat Rearrangement Challenge~\citep{habitatrearrangechallenge2022} and the AI2-THOR Rearrangement Challenge~\citep{RoomR}.
There is rich literature on object rearrangement in robotics~\citep{goodwin2022semantically,huang2019large,gu2022multi,cheong2020relocate,vasilopoulos2021reactive,liang2022code,zhang2022visually}. 
A common assumption in those methods is that a goal arrangement is part of the input, and the robot knows the exact desired positions of objects. 
ALFRED~\citep{shridhar2020alfred} proposed a language-based multi-step object rearrangement task, for which a number of solutions have been proposed that combine high-level skills~\citep{blukis2022persistent,min2021film}, and which have recently been extended to use LLMs as input~\citep{inoue2022prompter}. However, these operate at a very coarse, discrete level, instead of making motion-level and placement decisions, and thus can't make granular decisions about common-sense object arrangements.
By contrast, our work accepts underspecified instructions from humans, such as setting a dinner table with a few provided tableware objects.
\method{} has the capability to do common sense object rearrangement by extracting knowledge from LLMs, and operates both on a high level and on making motion-level placement decisions.

Object arrangement is a task that involves arranging items on a tabletop to achieve a specific functional, semantically valid goal configuration. 
This task requires not only the calculation of object positions but also adherence to common sense, such as placing forks to the left and knives to the right when setting a table. 
Previous studies in this area, such as~\citep{liu2022structformer,kapelyukh2022dall, liu2022structdiffusion, wei2023lego}, focused on predicting complex object arrangements based on vague instructions. 
For instance, StructFormer~\citep{liu2021structformer} is a transformer-based neural network for arranging objects into semantically specified structures based on natural-language instructions. 
By comparison, our approach \method{} utilizes an LLM for common sense acquisition to avoid the need of demonstration data for computing object positions. 
Additionally, we optimize the feasibility and efficiency of plans for placing tableware objects. 

There exist methods for predicting complex object arrangement using web-scale diffusion models~\citep{kapelyukh2022dall}. 
Their approach, called DALL-E-Bot, enables a robot to generate goal images using DALL-E~\citep{ramesh2022hierarchical} based on a text description derived from an initial scene. The robot aligns objects between the initial and generated images and accordingly arranges objects in a tabletop scenario based on the inferred poses. 
Similar to DALL-E-Bot, \method{} achieves zero-shot performance using pre-trained models, but it is not restricted to a single top-down view of a table. Additionally, \method{} leverages LLMs to generate symbolic and geometric goal specifications from underspecified instructions, operates in full-room mobile manipulation settings, and integrates task and motion planning (TAMP) to reason about feasibility and uncertainty in both navigation and manipulation, producing efficient and physically executable plans.

\subsection{Robot Planning with Foundation Models}
Many LLMs have been developed in recent years, such as BERT~\citep{devlin2018bert}, GPT-3~\citep{brown2020language}, ChatGPT~\citep{openai}, CodeX~\citep{chen2021evaluating}, and OPT~\citep{zhang2022opt}. 
These LLMs can encode a large amount of common sense~\citep{liu2021pre} and have been applied to robot task planning~\citep{kant2022housekeep,huang2022language,ahn2022can,huang2022inner,singh2022progprompt,ding2023integrating,liu2023llm+,zhao2023large,liu2022structdiffusion,wu2023tidybot,rana2023sayplan}. 
For instance, the work of Huang et. al. showed that LLMs can be used for task planning in household domains by iteratively augmenting prompts~\citep{huang2022language}. 
SayCan is another approach that enabled robot planning with affordance functions to account for action feasibility, where the service requests are specified in natural language (e.g., ``make breakfast'')~\citep{ahn2022can}. 
Compared with those methods, \method{} optimizes both feasibility and efficiency while computing semantically valid geometric configurations.

More recently, vision-language models (VLMs) have been incorporated into robot planning~\citep{huang2023voxposer,huang2024rekep,zhang2024dkprompt,yang2024guiding}. 
Among those, the most relevant is VLM-TAMP~\citep{yang2024guiding}, which is a hierarchical planning approach that
leverages a VLM to generate semantically-specified and intermediate subgoals that guide a task and
motion planner. 
VLM-TAMP used a VLM to unify the processing of both text and image prompts, and is strong in tabletop visual scene analysis compared with our work. 
Compared with their approach, ours is capable of visual understanding for MoMa behaviors, such as probabilistically evaluating their feasibility. 

Next, we present a statement of the MoMa problem (namely object rearrangement) in Section~\ref{sec:problem_statement}, including assumptions, formats of its input and output, and success criteria, before we discuss our approach in Section~\ref{sec:alg}.

\begin{figure*}
\begin{center}
    \includegraphics[width=\textwidth]{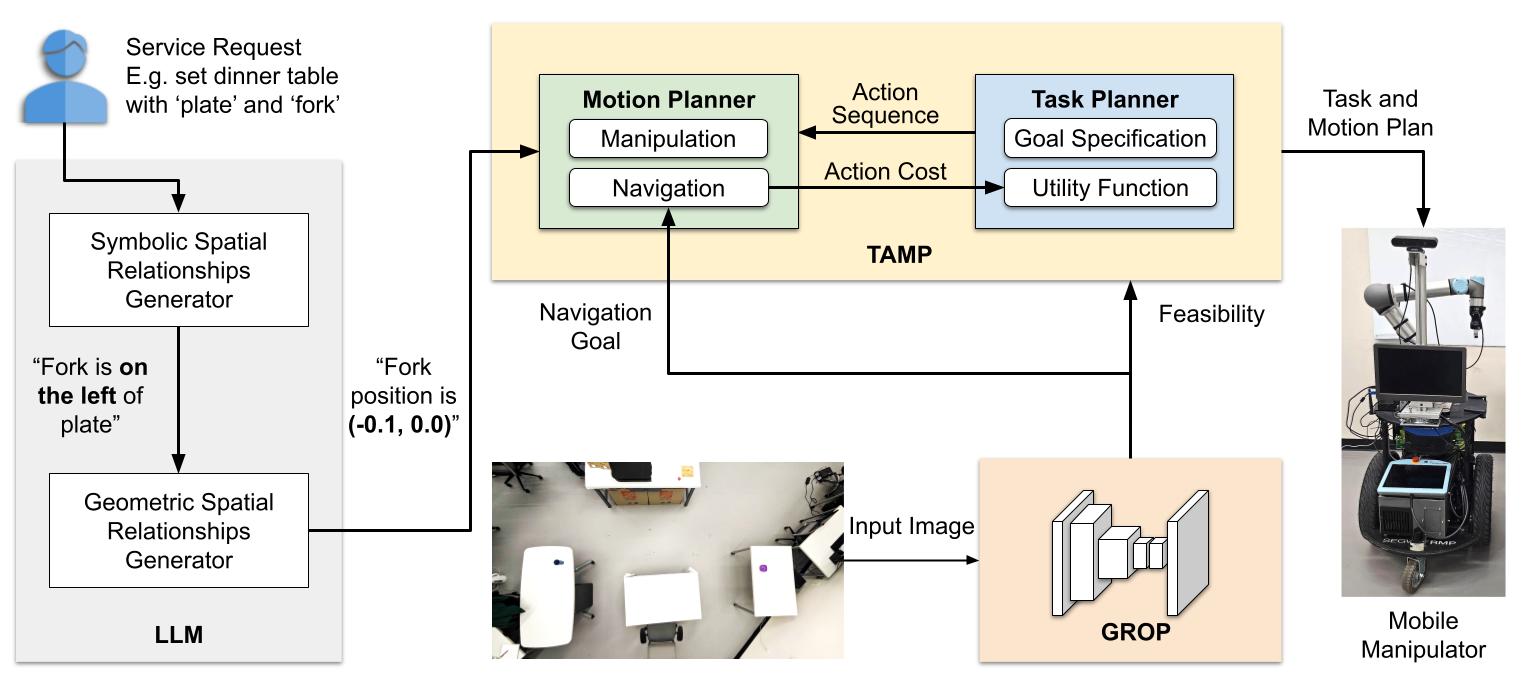} 
    \caption{
    An overview of the \method{} approach. 
    \method{} takes service requests from humans for setting tables and produces a task-motion plan that the robot can execute. 
    \method{} is comprised of two key components: the LLM and the Task and Motion Planner.
    The LLM is responsible for creating both symbolic and geometric spatial relationships between the tableware objects. 
    This provides the necessary context for the robot to understand how the objects should be arranged on the table.
    The Task and Motion Planner generates the 
    plan for the robot to execute based on the information provided by the LLM. An important component of \method{} is GROP that takes a top-down view image as the input and suggests standing positions to facilitate MoMa behaviors. 
    GROP is trained exclusively using simulation data. 
    In the real world, the robot estimates poses of objects and builds a digital twin for task and motion planning. 
    Details of GROP are shown in Figure~\ref{fig:overview_grop}.}
\label{fig:overview}
\end{center}
\end{figure*}

\section{Problem Statement}
\label{sec:problem_statement}
While we are generally concerned with a mobile manipulation domain, algorithms and systems developed in this article are demonstrated and evaluated using a specific task of object rearrangement. 
Specifically, the objective is to rearrange multiple tableware objects, 
which are initially scattered at different locations, into a tabletop configuration that is semantically valid and aligns with common sense.
The domain includes $N$ objects $\textit{Obj}$. 
There are obstacles (tables and chairs in our case) that prevent the robot from navigating to some positions in the domain. 
The robot is provided with prior knowledge about table shapes and locations.
Chairs, on the other hand, can only be sensed at planning time. 
Location $l$ is a symbolic concept that corresponds to a set of obstacle-free 2D poses ($X$), where each pose ($x\in X$) specifies a 2D position and an orientation. 
The robot needs to move each object $o \in \textit{Obj}$ from its initial location to a goal position. 

\vspace{.5em}
\noindent
\textbf{Actions: }The robot is equipped with skills of performing a set of symbolic (task-level) actions denoted as $A: A^n \cup A^m$, where $A^n$ and $A^m$ are \emph{navigation} actions and \emph{manipulation} actions respectively.
A navigation action $a^n_{l, l'} \in A^n$ is specified by its initial and goal locations, $l, l'\in L$, where $L$ includes a set of symbolic locations. 
A manipulation action, $a^m_{o, l} \in A^m$, is specified by an object to be manipulated, $o \in \textit{Obj}$, and a symbolic location, $l \in L$, to which the robot navigates and performs the manipulation action. 
We consider two types of manipulation actions of loading and unloading, represented by $a^{m+}$ and $a^{m-}$ respectively. 
Actions are defined via preconditions and effects. 
For instance, the action \texttt{load($o_1$)} has preconditions of \texttt{at(robot,$l_1$)} and \texttt{at($o_1$,$l_1$)}, meaning that to load the object $o_1$, the object must be co-located with the robot at the location $l_1$. 
The effects of \texttt{load($o_1$)} include $o_1$ being moved into the robot’s hand, i.e., \texttt{inhand($o_1$)}. 

\noindent
\textbf{Perception: }
The robot visually perceives the environment through top-down views over the areas where manipulation and navigation actions are performed. 
We use $\textit{IM}$ to represent a 2D image that captures the current obstacle configuration, such as chairs around tables, as shown in the ``Image Input'' of Figure~\ref{fig:overview} (bottom right). 
To facilitate robot learning, we provide a dataset (as illustrated in the ``Dataset'' box of Figure~\ref{fig:overview}). 
Each instance includes a top-down view image, and a target object with a predefined position, while each label is in the form of a heatmap. 
Each pixel of a heatmap is associated with a 2D position, and has a ``feasibility'' value that represents the success rate of the robot navigating to the 2D position, and manipulating the target object from there. 

A map is generated in a pre-processing step, and provided to the robot as prior information for navigation purposes using rangefinder sensors.

\noindent
\textbf{Uncertainty: }
We consider uncertainty in navigation and manipulation behaviors. 
For instance, the robot can fail in navigation  (at planning or execution time) when its goal is too close to tables or chairs, and it can fail in manipulation when it is not close enough to the target position. 
Note that uncertainties are treated as black boxes in this work.

Specifically, the outcome of performing navigation action $a^n_{l,l'}$ to goal pose $x$ is deterministic at the task level, but is non-deterministic at the motion level. 
In other words, the robot will end up in position $x'$, which is not necessarily the same as $x$. 
This setting captures the fact that a mobile robot never achieves its exact 2D navigation goal (due to its imperfect localization and actuation capabilities), though successfully navigating to an area ($l$) is generally possible. 

We focus on the interdependency between navigation and manipulation actions. 
For instance, the execution-time uncertainty from navigation actions results in different standing positions of the robot, which makes the outcomes of manipulation actions non-deterministic. 
This challenge generally exists in mobile manipulators. 
We assume no noise in the execution of manipulation actions (loading and unloading) to objects within a reachable area. 

\noindent
\textbf{Large Language Models:}
It is assumed that an off-the-shelf large language model (LLM) is available and can be used by a robot to extract common sense knowledge. 
An LLM is a type of computational model designed for natural language processing tasks such as language generation. 
The input of LLMs in this object rearrangement domain include a decription of the problem as well as the format of the output. 
It it evident that an LLM's performance can be boosted by providing a few inference examples~\citep{kaplan2020scaling}, so MoMa practitioners might want to extend the input by further including such examples.

\noindent
\textbf{Format of Solution: }
A solution is in the form of a task-motion plan 
$
    p=\langle p^t, p^m \rangle, 
$
where task plan $p^t$ is of the form $\langle a_0^n, a_0^m, a_1^n, a_1^m, ...\rangle$, indicating that navigation and manipulation actions are interleaved. 
Motion plan $p^m$ is of the form $\langle \xi_0^n,\xi_0^m, \xi_1^n,\xi_1^m, ... \rangle$, and $\xi_i^n$ (or $\xi_i^m$) is a trajectory in continuous space for implementing symbolic action $a_i^n$ (or $a_i^m$).

\noindent
\textbf{Quality of Solution: }
The quality of task-motion plan $p$ is evaluated using a utility function $\mathcal{U}(p)$, which considers both feasibility and efficiency of plan $p$:
\begin{align}
   \mathcal{U}(p) = \mathcal{R} \cdot \mathcal{F}(p) - \mathcal{C}(p),
\label{eqn:utility}
\end{align}
where $\mathcal{F}(p) \in [0,1]$ is the plan feasibility (i.e., the probability that $p$ can be successfully executed), $\mathcal{C}(p)$ is the overall plan cost of executing $p$, and $\mathcal{R}\!\!\!\rightarrow \!\!\!\mathbb{R}$ is a success bonus reflecting the reward from a successful execution. 
An optimal algorithm reports a task-motion plan of the highest utility: 
$$
    p^* = \argmax_{p} ~\mathcal{U}(p) 
$$

Next, we present \method{} that computes goal configurations of objects, and task-motion plans for realizing the goal, through visually grounding spatial relationships while considering both efficiency and feasibility. 

\section{Method}
\label{sec:alg}

In this article, we develop \method{}, a task and motion planning (TAMP) approach that is semantically specified and visually grounded, as applied to object rearrangement tasks. 
At the high level, \method{} generates object goal configurations (i.e., 2D positions) of relevant objects using common sense knowledge extracted from the LLM (Section~\ref{sec:llm}). 
At the low level, \method{} compute TAMP solutions for grounding the generated object locations into the physical world (Section~\ref{sec:grop}). 

\subsection{LLM-Guided Goal Generation}\label{sec:llm}

This subsection presents our approach that leverages LLMs to compute both symbolic spatial relationships (e.g., a fork should be placed to the left of a plate and a knife to the right) and geometric spatial relationships (e.g., 2D coordinates of the objects on a table). 

\noindent
\textbf{Generating Symbolic Spatial Relationships: }\label{sec:symbolic}
LLMs are first used to extract common sense knowledge regarding symbolic spatial relationships among objects placed on a table.
This is accomplished through the utilization of a template-based prompt:

\begin{quote}
Template 1: \emph{The goal is to set a dining table with objects. The symbolic spatial relationship between objects includes [spatial relationships]. [examples]. What is a typical way of positioning [objects] on a table? [notes].}
\end{quote}

\begin{figure*}[t]
\begin{center}
    \includegraphics[width=0.95\textwidth]{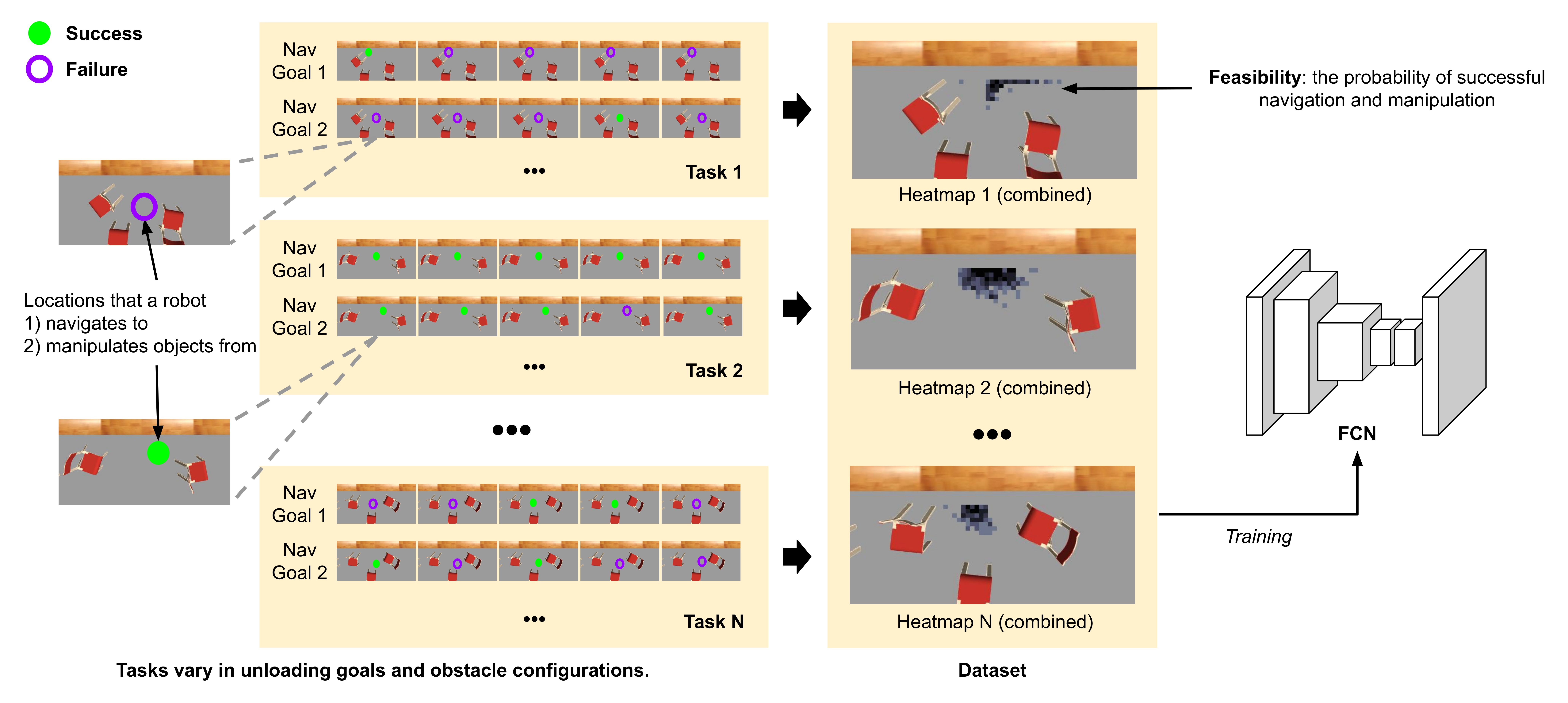}
    \caption{An overview of the data collection and training process in GROP.
    A \emph{task} corresponds to one ``unloading goal'' on the table, as well as a configuration of obstacles (chairs in our case). 
    Given a task, every pixel is considered a navigation goal -- the robot attempts to navigate there, and unload an object from there. 
    This navigation-manipulation process is referred to as a \emph{trial}. 
    The robot performs multiple trials for each navigation goal, which yields a \emph{feasibility} value for that particular location. 
    The feasibility values together form one \emph{heatmap} for each task. 
    In our \emph{dataset}, each instance is a top-down view image, whose label is the corresponding heatmap. 
    The ``Dataset'' box shows a few ``combined heatmaps'' where heatmaps are overlaid onto the corresponding images. 
    Training with the dataset generates an FCN that is used for two purposes: 1)~evaluating the feasibility of task-level actions, and 2) selecting motion-level navigation goals. 
    Finally, GROP incorporates both efficiency (measured by action costs) and feasibility to compute task-motion plans for a mobile manipulator. 
    } 
    \label{fig:overview_grop}
\end{center}
\end{figure*}

\noindent where \emph{[spatial relationships]} includes a few spatial relationships, such as \emph{to the left of} and \emph{on top of}.
In presence of \emph{[examples]}, the prompting becomes few-shot; when no examples are provided, it is simplified to zero-shot prompting. 
In practice, few-shot prompts can ensure that the LLM's response follows a predefined format, though more prompt engineering efforts are needed. 
\emph{[objects]} refers to the objects to be placed on the table, such as \emph{a plate}, \emph{a fork}, and \emph{knife}. 
To control the LLM's output, \emph{[notes]} can be added, such as the example \emph{``Each action should be on a separate line starting with `Place'. The answer cannot include other objects''}.

LLMs are generally reliable in demonstrating common sense, but there may be times when they produce contradictory results. 
To prevent \textbf{logical errors}, a logical reasoning-based approach has been developed to evaluate the consistency of generated candidates with explicit symbolic constraints.
This approach is implemented on answer set programming (ASP), which is a declarative programming language that expresses a problem as a set of logical rules and constraints~\citep{gebser2008user}. 
In the event of a logical inconsistency, the same template is repeatedly fed to the LLM in an attempt to elicit a different, logically consistent output.
ASP enables recursive reasoning, where rules and constraints can be defined in terms of other rules and constraints, providing a modular approach to problem-solving~\citep{jiang2019taska}. 
ASP is particularly useful for determining whether sets of rules and constraints are true or false in a given context.

The approach involves defining spatial relationships, their transitions, and rules for detecting conflicts.
These rules are created by human experts and serve to ensure that the generated context is logical and feasible.
One such rule is \verb|:- below(X,Y),right(X,Y)|, which states that object \verb|X| cannot be both ``below'' and ``to the right of'' object \verb|Y| at the same time.
This rule ensures that the resulting arrangement of objects is physically possible.
An instance of identifying a logical error is provided. 
For example, an LLM may generate instructions for arranging objects as follows:
\begin{enumerate}
    \item {Place fruit bowl in the center of table.}
    \item {\emph{Place {butter knife} above and to the right of {fruit bowl}}.}
    \item {\emph{Place {dinner fork} to the left of {butter knife}}.}
    \item {Place dinner knife to the right of butter knife.}
    \item {\emph{Place {fruit bowl} to the right of {dinner fork}}.}
    \item {Place water cup below and to the left of dinner knife.}
\end{enumerate}

Objects and spatial relationships are explicitly listed in the prompts used for querying LLMs, as shown in Template 1. 
We then use standard search methods to extract the objects and spatial relationships from the LLM outputs. 
In most cases, the LLMs are able to output instructions in the desired format; otherwise, we re-prompt the LLMs until we are able to extract the objects and spatial relationships.

There are \textbf{logical inconsistencies} in the italic lines: Steps 2 and 3 suggest placing the \emph{fruit bowl} below the \emph{dinner fork}, while Step 5 suggests placing the \emph{fruit bowl} to the right of the \emph{dinner fork}.
This contradicts the established rule and results in no feasible solutions.

\noindent
\textbf{Generating Geometric Spatial Relationships}~\label{sec:geometric}
After determining the symbolic spatial relationships between objects, we move on to generate their geometric configurations, where we use the following LLM template.

\begin{quote}
Template 2: \emph{[object A] is placed [spatial relationship] [object B]. How many centimeters [spatial relationship] [object B] should [object A] be placed?}
\end{quote}

For instance, when we use Template 2 to generate prompt ``\emph{A dinner plate is placed to the left of a knife. How many centimeters to the left of the water cup should the bread plate be placed?}'', GPT-3 produces the output ``\emph{Generally, the dinner knife should be placed about 5-7 centimeters to the right of the dinner plate.}''
In practice, the exact distance is extracted by searching for keyword ``centimeter'' to identify the number preceding it. 

To determine the positions of objects, we first choose a coordinate origin.
This origin could be an object that has a clear spatial relationship to the tabletop and is located centrally. 
A dinner plate is a good example of such an object.
We then use the recommended distances and the spatial relationships between the objects to determine the coordinates of the other objects. 
Specifically, we can calculate the coordinates of an object by adding or subtracting the recommended distances in the horizontal and vertical directions, respectively, from the coordinates of the coordinate origin.
The LLM-guided position for the $i$th object is denoted as $(x^i, y^i)$, where $i\in N$.

However, relying solely on the response of the LLMs is not practical as they do not account for object attributes such as shape and size, including tables constraints.
To address this limitation, we have designed an adaptive sampling-based method that incorporates object attributes after obtaining the recommended object positions.
Specifically, our approach involves sequencing the sampling of each object's position using a 2D Gaussian sampling technique~\citep{boor1999gaussian}, with $(x^i, y^i)$ as the mean vector, and the covariance matrix describing the probability density function's shape.

The resulting distribution is an ellipse centered at $(x^i, y^i)$ with the major and minor axes determined by the covariance matrix. 
However, we do not blindly accept all of the sampling results; instead, we apply multiple rules to determine their acceptability, inspired by rejection sampling~\citep{gilks1992adaptive}.
These rules include verifying that the sampled geometric positions adhere to symbolic relationships at a high level, avoiding object overlap, and ensuring that objects remain within the table boundary. 
For example, if the bounding box of an object position falls outside the detected table bounds, we reject that sample.
The bounding box of objects and the table are computed based on their respective properties, such as size or shape.
After multiple rounds of sampling, we can obtain $M$ object configuration sequences.

The output of our LLM-guided goal generation approach is a 2D tabletop configuration of relevant objects. 
Next, we describe our visually grounded TAMP approach for realizing the computed goal configurations.

\begin{figure*}
\begin{center}
    \includegraphics[width=0.87\textwidth]{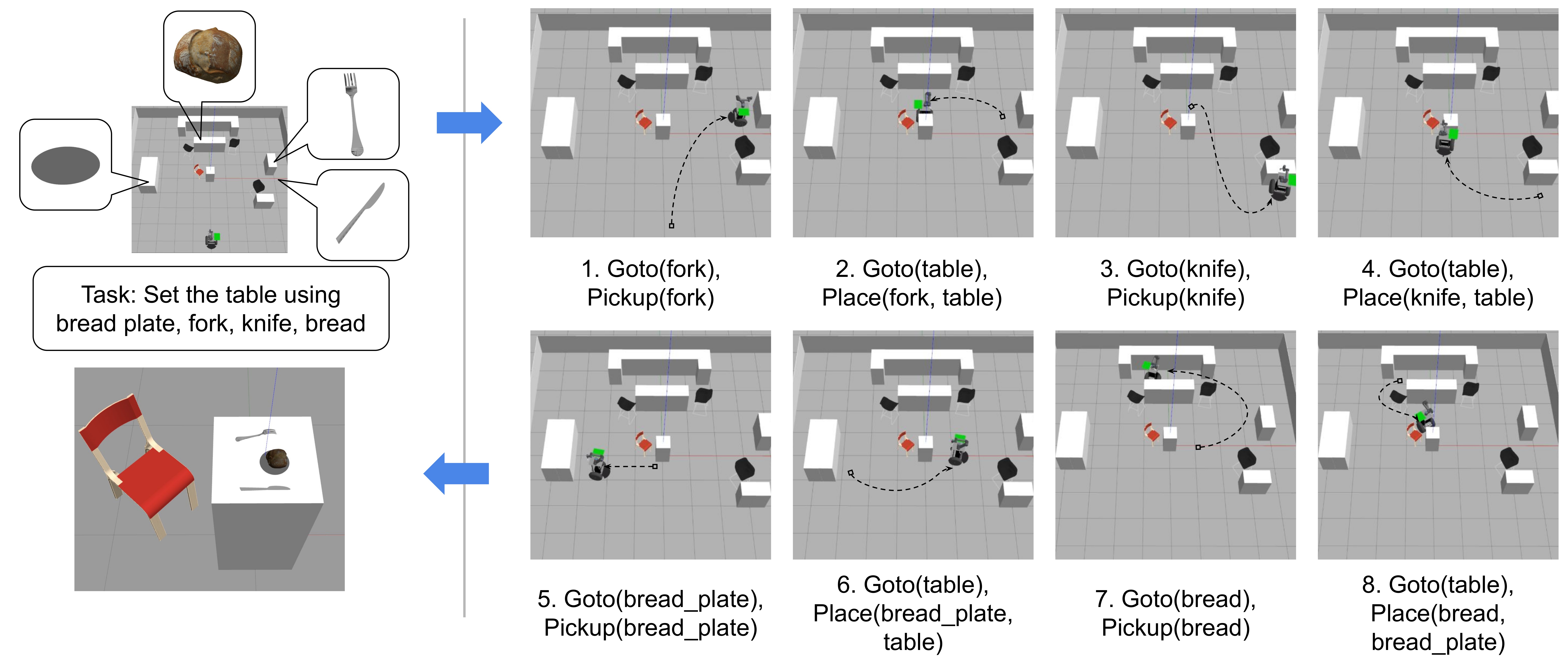}
    \caption{An illustrative example of \method{} showing the robot navigation trajectories (dashed lines) as applied to the task of ``set the table with a bread plate, a fork, a knife, and a bread.'' \method{} is able to adapt to complex environments, using common sense extracted from an LLM to generate efficient (i.e., minimize the overall navigation cost) and feasible (i.e., select an available side of the table to unload) pick-and-place motion plans for the robot.
    }
    \label{fig:illustrative_example}
\end{center}
\end{figure*}

\subsection{Task and Motion Planning for Grounded Object Rearrangement}\label{sec:grop}
After identifying feasible object configurations on the tabletop, the next step is to place the objects on the tabletop based on one of object configuration sequences. 
At the task level, the robot must decide the sequence of object placement and how to approach the table. 
For example, if a bread is on top of a plate, the robot must first place the plate and then the bread. 
The robot must also determine how to approach the table, such as from which side of the table.
Once the task plan is determined, the robot must compute 2D navigation goals (denoted as $loc$) at the motion level that connect the task and motion levels. 
Subsequently, the robot plans motion trajectories for navigation and manipulation behaviors. 

In the presence of dynamic obstacles, not all navigation goals ($loc$) are equally preferred. 
For instance, it might be preferable for the robot to position itself close to an object for placement rather than standing at a distance and extending its reach. 
As a result, we developed a novel TAMP component (namely GROP) in our system for computing the optimal navigation goal $loc$, which enabled the task-motion plan with the maximal utility for placing each object in terms of feasibility and efficiency given an object configuration $(x^i_j, y^i_j)$, where $0\leq j\leq M$.

\noindent
\textbf{The GROP Algorithm:}
Algorithm~\ref{alg:GROP} presents the GROP algorithm. 
Implementing GROP requires a task planner $\textit{Plnr}^t$, a motion planner $\textit{Plnr}^m$ for planning base motions, a success bonus $\mathcal{R}\!\!\rightarrow \!\!\mathbb{R}$, and a cost function $\textit{Cst}$ that evaluates the cost of any motion trajectory generated by $\textit{Plnr}^m$. 
Inputs of GROP include a rule-based task description $T$, a robot initial 2D position $x^{\textit{init}}$, and a provided dataset $D$.
GROP outputs a task-motion plan $p$ in the form of $\langle p^t, p^m \rangle$.

GROP starts with training an FCN-based feasibility evaluator $\Psi$ using provided dataset $D$ in Line~\ref{l:train}. 
Then it initializes an empty set of task-motion plans $\textbf{P}$ in Line~\ref{l:initP}.
$\textit{Plnr}^t$ takes $T$ as input and outputs a set of task-level satisficing plans, denoted as $\textbf{P}^t$ in Line~\ref{l:task}.
The outer for-loop (Lines~\ref{l:oloops}-\ref{l:oloope}) iterates over each task-level satisficing plan.
In each iteration, GROP evaluates the utility value of one task plan $\mathcal{U}(p)$, which incorporates both plan feasibility $\mathcal{F}(p)$ and plan efficiency $\mathcal{C}(p)$.
Aiming to evaluate $\mathcal{F}(p)$ and $\mathcal{C}(p)$, each iteration in the first inner for-loop (Lines~\ref{l:iloops}-\ref{l:iloope}) considers a pair of navigation and manipulation actions in the task plan, and evaluates its feasibility and cost.
In the second inner for-loop of Lines~\ref{l:motions}-\ref{l:motione}, GROP calls $\textit{Plnr}^m$ to compute one motion plan for each task-level action.
While the generated motion plan $p^m$ only includes base motion, the feasibility of arm motion (Eqn.~\ref{eqn:task_feasibility} and Line~\ref{l:feas}) is considered in our motion plan generation. 
Line~\ref{l:tmp} puts together task plan $p^t$ and motion plan $p^m$ to form a task-motion plan $p$.
In the same line, $p$ is added to task-motion plan set $\textbf{P}$.
Lines~\ref{l:select}-\ref{l:return} are the final steps to select and return the optimal task-motion plan from $\textbf{P}$ given utility function $\mathcal{U}(p)$.

For different groups of object configurations, we use GROP to compute the maximal utility value of task-motion plans and select the best one for execution.
Figure~\ref{fig:illustrative_example} shows one task-motion plan generated using \method{} for a four-object rearrangement task.

\begin{algorithm}[t]
\caption{GROP}\label{alg:GROP}  \small
\begin{algorithmic}[1]
\REQUIRE Task planner $\textit{Plnr}^t$, motion planner $\textit{Plnr}^m$, success bonus $\mathcal{R}$, and cost function $\textit{Cst}$\\
\hspace{-1.7em}\textbf{Input:} Task description $T$, robot initial position $x^{\textit{init}}$, dataset $D$
\STATE {Train a motion-level feasibility evaluator $\Psi$ using dataset $D$ (detailed in Figure~\ref{fig:overview_grop})} \label{l:train}
\STATE {Initialize a set of task-motion plans $\textbf{P} \leftarrow \emptyset$} \label{l:initP}
\STATE {Compute a set of task-level satisficing plans: $\textbf{P}^t\leftarrow \textit{Plnr}^t(T)$} \label{l:task}
\FOR {each plan $p^t \in \textbf{P}^t$} \label{l:oloops}
    \STATE {Initialize a motion-level position sequence: $X^{\textit{seq}} \leftarrow [x^{\textit{init}}]$}
    \STATE {Initialize $\textit{tmp}^f \leftarrow 0$ and $\textit{tmp}^c \leftarrow 0$}
    \FOR {each action pair $\langle a^n_{l, l'}, a^m_{o, l'} \rangle$ in $p^t$} \label{l:iloops}
        \STATE {Capture $\textit{IM}$ of location $l'$}
        \STATE {Predict heatmap $h = \Psi(\textit{IM})$, using Eqn.~\ref{eqn:heatmap}}
        \STATE {$\textit{tmp}^f \leftarrow \textit{tmp}^f + \textit{Fea}^t(a^n_{l, l'}, a^m_{o, l'} )$, using Eqn.~\ref{eqn:task_feasibility}} \label{l:feas}
        \STATE {$x' \leftarrow \textit{Smp}( l',h )$, and append $x'$ to $X^{\textit{seq}}$} \label{l:smp}
        \STATE {$\textit{tmp}^c \!\! \leftarrow \!\! \textit{tmp}^c + \textit{Cst}\big(\textit{Plnr}^m(a^n_{l, l'})\big) + \textit{Cst}\big(\textit{Plnr}^m(a^m_{o, l'})\big)$ }
    \ENDFOR \label{l:iloope}
    \FOR{each $(x_i, x_{i+1}) \in X^{\textit{seq}}$} \label{l:motions}
        \STATE {Compute motion-level trajectory $\xi \leftarrow \textit{Plnr}^m(x_i, x_{i+1})$}
        \STATE {Append $\xi$ to motion plan $p^m$}
    \ENDFOR \label{l:motione}
    \STATE {Generate task-motion plan $p \leftarrow \langle p^t, p^m \rangle$, and append $p$ to the task-motion plan set $\textbf{P}$} \label{l:tmp}
    \STATE Update $\displaystyle \mathcal{F}(p)\leftarrow \frac{\textit{tmp}^f}{\lvert p^t\rvert}$
       and $\mathcal{C}(p)\leftarrow \textit{tmp}^c$
    \STATE {$\mathcal{U}(p) \leftarrow \mathcal{R} \cdot \mathcal{F}(p) - \mathcal{C}(p)$ (Eqn.~\ref{eqn:utility})}
\ENDFOR \label{l:oloope}

\STATE {Compute optimal task-motion plan: $p^* = \argmax_{p \in \textbf{P}} \mathcal{U}(p)$} \label{l:select}

\RETURN {$p^*$} \label{l:return}

\end{algorithmic}
\end{algorithm}

\noindent
\textbf{Motion-Level Feasibility Evaluation in GROP: }
In our mobile manipulation domain, motion-level feasibility $\textit{Fea}^m(x,y)$ is a function of 2D positions $x$ and $y$, and is the probability of a robot successfully navigating to $x$ and manipulating an object that is in position $y$. 
$\textit{Fea}^m(x, y)$ can be extracted from gray-scale heatmap image $h^y$ that is centered around $y$: 
\begin{align}
    \textit{Fea}^m(x,y) = h^y[x]
\end{align}

We use a FCN-based feasibility evaluator $\Psi$ to generate heatmap $h^y$, given a top-down view image $\textit{IM}^y$ captured right above unloading position $y$ (``Image Input'' in Figure~\ref{fig:overview}): 
\begin{align}
    h^y = \Psi(\textit{IM}^y)
\label{eqn:heatmap}
\end{align}

\noindent
\textbf{Data Collection and Learning $\Psi$ with FCN: }
Here we discuss how to learn $\Psi$ in Equation~\ref{eqn:heatmap}. 
A \emph{task} specifies an obstacle configuration and a position $y$ that a robot wants to unload objects to. 
In each \emph{trial} of our data collection process, a robot attempts to navigate to position $x$, and then unload an object to position $y$. 
Such a trial produces a data point in the following format: 
$$
    (\textit{IM}^y, x):r
$$
where $\textit{IM}^y$ is a top-down view image captured right above $y$, and $r$ is either $\textit{true}$ or $\textit{false}$ depending on if the robot succeeds in both navigation and manipulation actions. 
The robot repeated the same process for $N$ times ($N=5$ in our case), and we used the results ($r_0, r_1, \cdots, r_{N-1}$) to compute a success rate for positions $x$ and $y$, which determines a gray-scale color for one pixel of a heatmap: $h[x]$.

Iterating over all possible positions of $x$ in an area of $\textit{Width} \times \textit{Height}$ ($24$ pixels by $8$ pixels in our case) in image $\textit{IM}$, we were able to generate one full heatmap $h$ for the current task. 
Here we assume this area is large enough to cover all positions, from which the robot can unload objects to $y$. 
To diversify the instances, we randomly placed obstacles (chairs in our case) to generate ten different ``environments,'' and then randomly sampled unloading positions to generate a total of 100 tasks. 
As a result, our dataset contains 100 instances, each in the form of a top-down view image ($64\times 32$). 
Each instance has a label that is in the form of a heatmap. 
The size of our dataset is $96,000$, i.e., $100\! \times \! N \! \times \! \textit{Width}\!  \times \! \textit{Height}$. 
The data collection and learning process is illustrated in Figure~\ref{fig:overview_grop}. 

Our feasibility evaluator, $\Psi$, is trained exclusively on simulated data. 
We train the FCN as a N-class pixel-level classifier using softmax and a standard cross-entropy loss on the discrete success-level labels. At inference time, rather than selecting the top class, we convert the logits into class probabilities and take a weighted average over the $N$ levels, and this produces smooth, continuous heatmaps as shown in Figure~\ref{fig:overview_grop}.

To deploy in the real world, we first estimate the poses of obstacles (i.e., tables and chairs) and reconstruct a digital twin of the workspace but using the same simulated assets from our training phase. We then render top-down views as the RGB observations and feed them directly into $\Psi$, sidestepping the sim-to-real gap.
We train the FCN as a N-class pixel-level classifier using softmax and a standard cross-entropy loss on the discrete success-level labels. At inference time, rather than selecting the top class, we convert the logits into class probabilities and take a weighted average over the N levels, and this produces smooth, continuous heatmaps as shown in Figure 3.

For real-world experiments (e.g., the top-down view image in Figure~\ref{fig:overview}), the visual input can capture a large area with multiple tables. 
We only consider feasibility evaluation on the table for placing the objects. 
There are other tables from which the robot needs to retrieve objects, where we assume there is sufficient free space for picking up the objects and we do not analyze motion feasibility for those tables. 

The current training data was collected when the robot moves objects between rectangle-shaped tables, and the objects are distant from each other on the tables. 
If the table is of irregular shape or the tabletop objects are close to each other, the rearrangement tasks become more difficult and it is likely LLM-GROP will not work well. 
Otherwise, we expect that LLM-GROP’s performance will be robust to object positions on the table and shapes of obstacles. Manipulation behaviors other than grasping and ungrasping, e.g., pouring water and folding cloth, can be equipped onto the robot as long as additional training data is provided to guide the robot in selecting standing positions.

\noindent
\textbf{Task-Level Feasibility Evaluation in GROP: }
$\textit{Fea}^t(a^n_{l, l'}, a^m_{o, l'})$ evaluates the feasibility (in the form of a probability) of a robot successfully performing both task-level navigation action $a^n_{l, l'}$ and task-level manipulation action $a^m_{o, l'}$. 
\begin{align}
    \textit{Fea}^t(a^n_{l, l'}, a^m_{o, l'}) = \frac {\sum\limits_{i=0\cdots N-1} \!\!\! \textit{Fea}^m\big( \textit{Smp}_i(l',h), y \big) }{N}
\label{eqn:task_feasibility}
\end{align}
where function $\textit{Smp}_i(l', h)$ samples the $i$th 2D position from location $l'$. The positions are weighted by heatmap $h$ that is centered around object $o$. 
Intuitively, positions of higher motion-level feasibility are more likely to be sampled.

This section presents the two key components of \method{} for 1) computing semantically specified goal configurations of objects using common sense extracted from LLMs and 2) visually grounded planning at both task and motion levels for realizing the goals.

\section{Experiments}
To evaluate the effectiveness of \method{}, we conducted a series of experiments focused on tableware object rearrangement tasks.
In these tasks, a mobile manipulator is assigned the job of setting a dinner table using a specific set of objects.
The experiments include nine distinct tasks involving various objects, as detailed in Table~\ref{table:task}.
The robot must retrieve multiple objects from different locations and place them on a central table, with obstacles (e.g., a chair) randomly positioned around the table.
These tasks require the robot to compute semantically valid tabletop arrangements, plan efficient object rearrangement strategies, and execute the plans through navigation and manipulation behaviors.

\begin{table}[t]
\scriptsize
\centering
\caption{Objects that are involved in our object rearrangement tasks for evaluation, where tasks 1-5 include three objects, tasks 6 and 7 include four objects, and task 8 and 9 includes five objects.}\label{table:task}
\begin{tabular}[t]{@{}cl@{}}
\toprule
	 Task \#ID & Objects\\ \midrule
	 1 & Dinner Plate, Dinner Fork, Dinner Knife\\ \midrule
      2 & Bread Plate, Water Cup, Bread\\ \midrule
	 3 & Mug, Bread Plate, Mug Mat\\ \midrule
      4 & Fruit Bowl, Mug, Strawberry\\ \midrule
      5 & Mug, Dinner plate, Mug Lid\\ \midrule\midrule
      6 & Dinner Plate, Dinner Fork, Mug, Mug Lid\\ \midrule
      7 & Dinner Plate, Dinner Fork, Dinner Knife, Strawberry\\ \midrule\midrule
      8 (Sim. Only) & Dinner Plate, Dinner Fork, Dinner Knife, Mug, Mug Lid\\ \midrule
      9 (Real Only) & Dinner Plate, Dinner Fork, Dinner Knife, Water Cup, Strawberry\\ 
\bottomrule
\end{tabular}
\end{table}

\subsection{Real Robot Experiment}
We begin with a real-world demonstration to validate the proposed approach.
The mobile manipulator used in this demonstration features a wheeled base for navigation and a 6-DOF robotic arm for performing manipulation tasks.

\begin{figure*}
  \begin{center}
    \includegraphics[width=.9\textwidth]{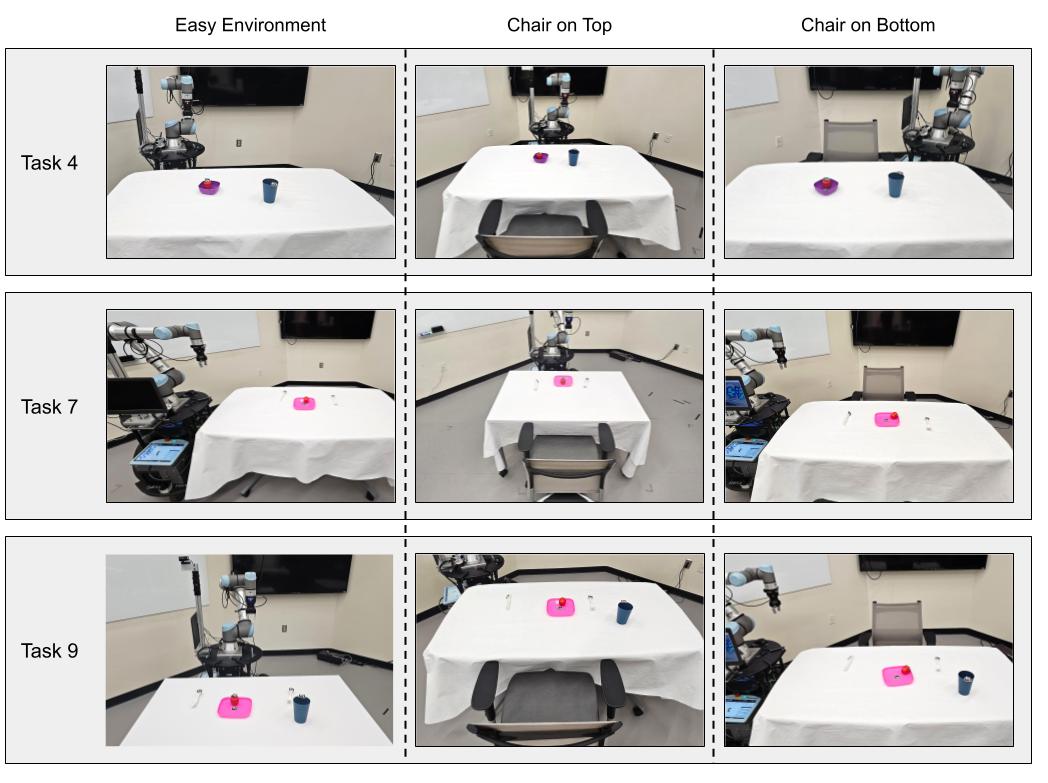}
    \caption{Example outcomes of the robot completing object rearrangement tasks. 
    The ``easy'' environment did not include any obstacles, while the other environments included a chair on one side of the table. 
    Note that the ``top'' and ``bottom'' labels shown in the columns were with respect to the robot's view. 
    There were three tasks (IDs 4, 7 and 9 -- see Table~\ref{table:task}) used for the real-robot experiments covering different numbers of objects being rearranged. 
    The robot dynamically computed the goal configurations of those objects and (task and motion) plans for realizing those configurations. 
    }
    \label{fig:trials_result}
  \end{center}
\end{figure*}

\noindent
\textbf{Experimental Setup:} 
We designed our real-world experiment to demonstrate that \method{} can effectively handle a variety of scenarios.
Tasks 4, 7, and 9 from Table~\ref{table:task} were selected for these demonstrations.
The environment consisted of three tables, with objects initially placed on the left and right tables, and the robot positioned randomly.
The robot repeated each task 15 times: during the first five repetitions, no obstacles were present; in the next five, a chair was positioned on the upper side of the table; and in the final five, a chair was positioned on the lower side of the table.

In the tableware object rearrangement tasks, task success was determined by whether the final positions of the objects appeared natural.
To evaluate this, we captured an overhead image of the table at the end of each trial and assessed whether the object arrangement looked logically appropriate or exhibited any unnatural placements.
To assess the effectiveness of \method{}, we recruited ten graduate students to evaluate the robot's performance by reviewing the captured images.
We implemented a five-point rating system, detailed in Table~\ref{table:criteria}, and asked the volunteers to score the tableware arrangements in the images provided.

We use the MoveIt software~\citep{coleman2014reducing} for planning grasping and ungrasping behaviors on the real robot, and there was no machine learning involved. 
To further facilitate robot grasping, we used QR codes to help the robot locate the objects of interest. 
For those objects that are hard to pick up, such as forks and knives, we use a utensil holder to improve the success rate of object grasping. 

For our real-robot experiments, we utilized OpenAI's GPT-3 engines as the LLM component.
The specific hyperparameters used in the study are provided in Table~\ref{table:parameter}.
While there exist newer LLMs, we decided to use GPT-3 as the default LLM to be consistent to the results reported in the two preceding conference papers~\citep{ding2023task,zhang2022visually}. When we made comparisons between LLM-GROP and baselines, it was strictly enforced that the same LLM was used in all methods to avoid the potential bias introduced by the LLMs. 
In addition, we report results comparing multiple versions of LLM-GROP implemented using different LLMs in Section 5.3, to demonstrate that LLM-GROP is capable of adapting to newer LLMs.

\begin{table}[t]
\caption{Rating guidelines for human raters in the experiments. \textbf{1} point indicates the poorest tableware object arrangement as it suggests that some objects are missing. Conversely, \textbf{5} points represent the best arrangement.}
\label{table:criteria}
\begin{threeparttable}
\scriptsize
\begin{tabular}{cl}
\toprule
	 Points & Rating Guidelines\\ \midrule
	 1 & Missing critical items compared with the objects listed at the top of \\
      (Poor)  & the interface (e.g., dinner plate, dinner fork, dinner knife), making it \\
        & hardly possible to complete a meal.\\ \midrule
      2 & All items are present, but the arrangement is poor and major \\
        & adjustments are needed to improve the quality to a satisfactory level.\\ \midrule
	 3 & All items are present and arranged fairly well, but still there is\\
        & significant room to improve its quality.\\ \midrule
      4 & All items are present and arranged neatly, though an experienced\\
        & human waiter might want to make minor adjustments to improve.\\ \midrule
      5 & All items are present and arranged very neatly, meeting the aesthetic \\
      (Strong)  & standards of an experienced human waiter.\\
\bottomrule
\end{tabular}
\end{threeparttable}
\end{table}

\begin{figure*}
  \begin{center}
    \includegraphics[width=\textwidth]{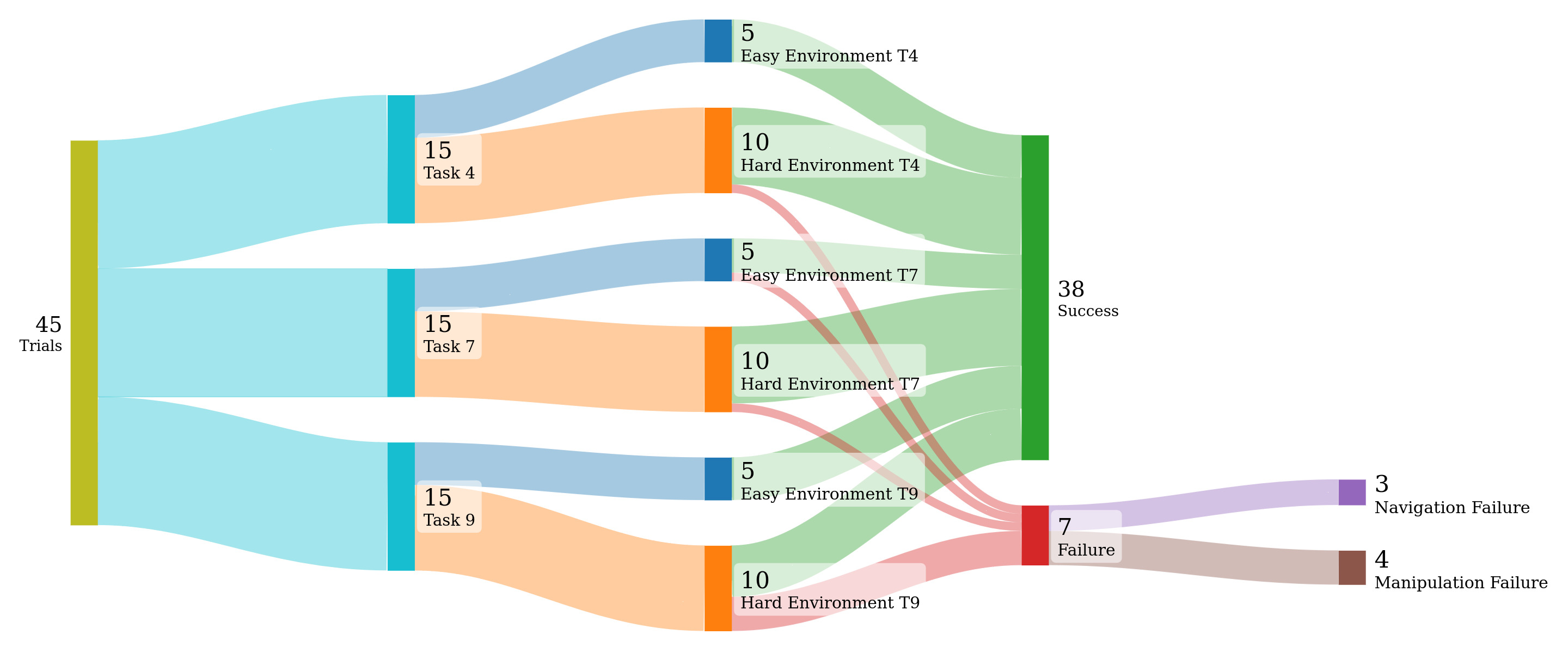}
    \caption{An overview of real robot experiment trials. There are 45 trials categorized in three tasks, and executed in two environments, an easy environment where there are no obstacles, and a hard one with a chair (obstacle) placed on top or bottom of the table. Successes total to 38 trials, while failure accounts to 7 trials, producing an overall 84.4\% success rate. The failure trials are further categorized based on the reason of failure, navigation failure (3 trials) and manipulation failure (4 trials). This graph highlights how task and environment complexity impact the success of each trial. 
    }
    \label{fig:sankey_graph}
  \end{center}
\end{figure*}

\noindent
\textbf{Demonstrations:} 
Figure~\ref{fig:trials_result} presents samples of the images collected during the real-robot demonstrations.
The GROP algorithm enables the robot to adapt its position based on variations in the chair's placement, ensuring correct object positioning regardless of the chair's location.
Simultaneously, the LLM component generates reasonable configurations for each task, maintaining accurate object placement even under changing conditions.
The results of the user ratings are summarized in Table~\ref{table:real robot exp}.
Performance decreases as the number of objects increases, and in more complex environments with obstacles, performance further declines due to the additional challenges of navigating around obstacles to reach the goal positions.
These results demonstrate the robustness of our approach in enabling a robotic platform to effectively perform real-world tasks.
To provide a detailed view of the outcomes, Figure~\ref{fig:sankey_graph} categorizes the 45 robot execution trials by task, environment complexity, success rate, and failure modes.

\begin{table}[t]
\caption{Hypermeters of OpenAI's GPT-3 engines in Our Experiment}
\label{table:parameter}
\scriptsize
\centering
\begin{tabular}{l|l||l|l}
\toprule
\textbf{Parameter} & \textbf{Value} & \textbf{Parameter} & \textbf{Value} \\
\midrule
Model & text-davinci-003 & Temperature & 0.1 \\
\midrule
Top p & 1.0 & Maximum length & 512 \\
\midrule
Frequency penalty & 0.0 & Presence penalty & 0.0 \\
\bottomrule
\end{tabular}
\end{table}

\begin{table}[t]
\centering
\caption{Average of user ratings score for three individual object rearrangement tasks within 3 environments, an easy one without any obstacles, and a hard one with the chair being placed to the top of the table and to the bottom of the table.}
\label{table:real robot exp}
\scriptsize
\begin{tabular}{c|l|l|l|l}
\toprule
	  \textbf{Task}  & \textbf{Easy Env.} & \parbox[t]{2cm}{\textbf{Hard Env. \\ (Chair to \\ the top)}}
 & \parbox[t]{2cm}{\textbf{Hard Env. \\ (Chair to \\ the bottom)}}
\\ 
    \midrule
	 Task 4 & \textbf{3.80} & 3.60 & 3.73\\
    \midrule
  Task 7 & 3.53 & 3.73 & \textbf{3.93} \\ 
    \midrule
	 Task 9 & \textbf{3.73} & 3.13 & 3.06  \\ 
 
\bottomrule
\end{tabular}
\end{table}

\subsection{Performance Assessment}
To assess the overall performance of the method and gain insights into the contributions of \method{}, we conducted performance evaluations in simulated scenarios.

\noindent
\textbf{Experimental Setup:} 
As with the real-robot experiment's configuration, we we conducted eight different tasks in a simulated environment\footnote{Implemented in the Gazebo simulator}, as detailed in Table~\ref{table:task}.
We execute each task 20 times using the \method{} system with the same prompt templates described in Section~\ref{sec:llm}, and after each task is completed, we capture an image of the table, the chair, and the objects on the tabletop for later human evaluation.
We used the same LLM as the real-robot demonstration to carry out the simulation experiment.

\noindent
\textbf{Baselines:}
\method{} is evaluated by comparing its performance to three baselines, where the first baseline is the weakest. 

\begin{itemize}
    \item Task Planning with Random Arrangement (TPRA): 
    This baseline uses a task planner to sequence navigation and manipulation behaviors, while it randomly selects standing positions next to the target table and randomly places objects in no-collision positions on the table. 
    \item LLM-based Arrangement and Task Planning (LATP): It can predict object arrangements using LLMs and perform task planning. It uniformly samples standing positions around the table for manipulating objects. 
    \item GROP: It considers plan efficiency and feasibility for task-motion planning, and lacks the capability of computing semantically valid arrangements. Similar to TPRA, GROP also randomly places objects in no-collision positions on the table.
\end{itemize}

\begin{figure}[t]
\begin{center}
    \includegraphics[width=0.48\textwidth]{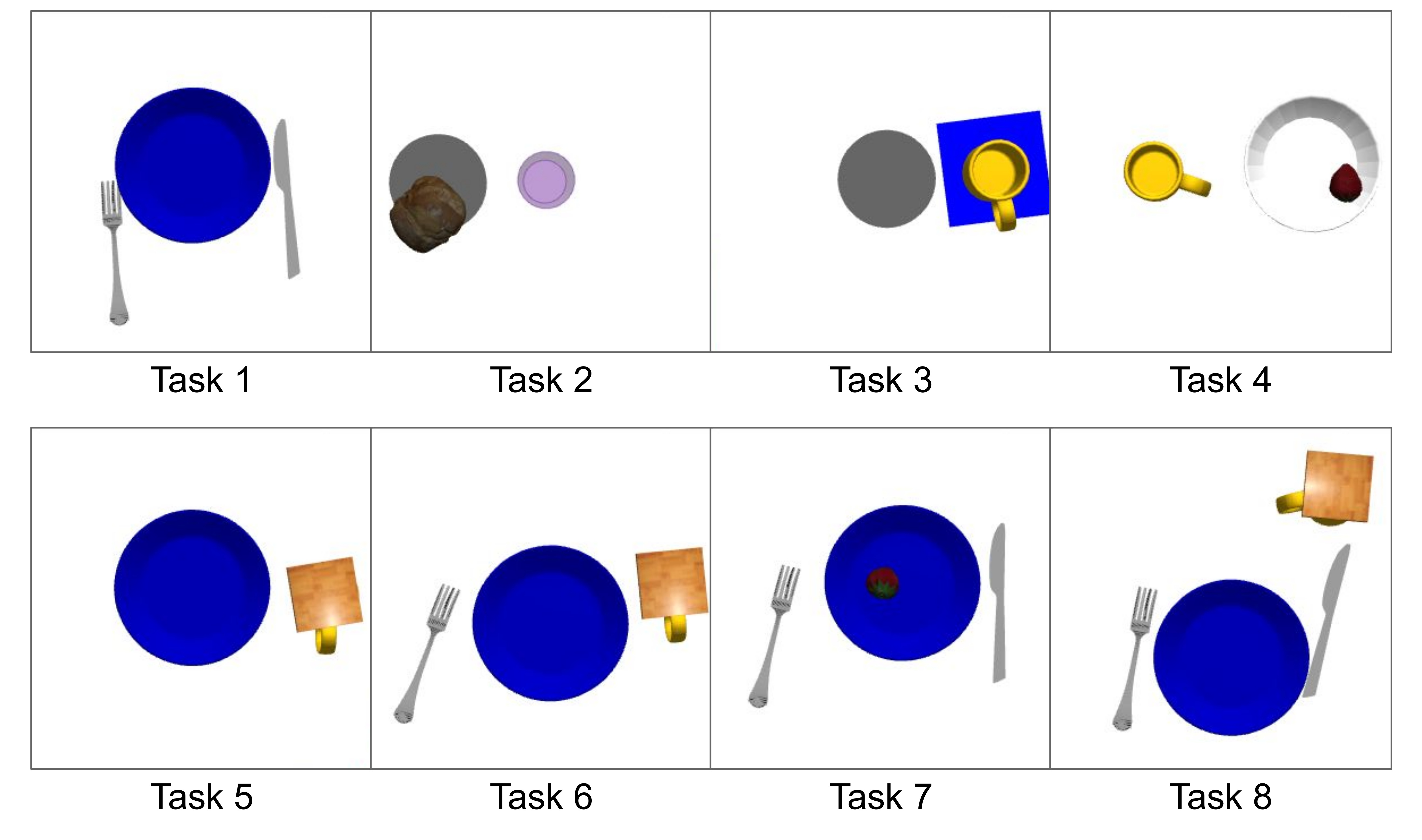}
    \caption{Examples of tableware objects rearranged by our \method{} agent in eight tasks, where the objects used in these tasks can be found in Table~\ref{table:task}.
    Our \method{} enables the arrangement of tableware objects to be both semantically valid.
    }\label{fig:8examples}
\end{center}
\end{figure}

\noindent
\textbf{Rating Criteria:}
We recruited five graduate students with engineering backgrounds, three females and two males between the ages of 22 and 30. 
We generated 640 images from the four methods (three baselines and \method{}) for eight tasks and each image required evaluation from all volunteers, resulting in a total sample size of 3200 images (the examples are shown in Figure~\ref{fig:8examples}). 
The volunteers were shown one image at a time on a website page that we provided, and they scored each image from 1 to 5 based on the rating rules.
We ensured that the rating was rigorous by using a website to collect rating results, thereby minimizing any potential biases that could arise from further interaction with the volunteers once they entered the website.

\begin{figure}
\begin{center}
    \includegraphics[width=0.45\textwidth]{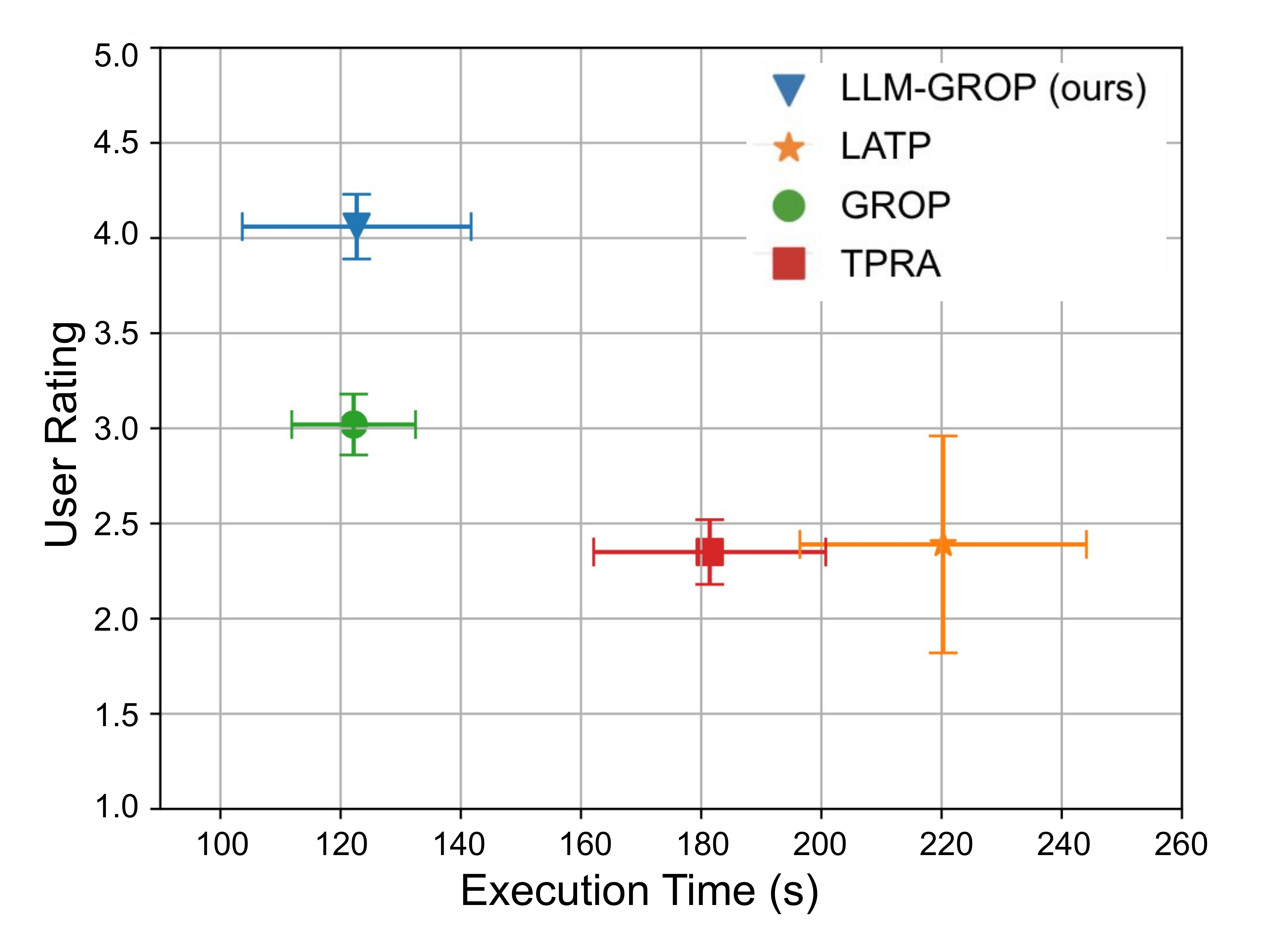}
    \caption{Overall performance of \method{} as compared to three baselines based on mean values and standard errors of user ratings and robot execution time for all tableware object arrangement tasks. 
    }\label{fig:main}
\end{center}
\end{figure}

\noindent
\textbf{\method{} vs. Baselines:} 
Figure~\ref{fig:main} shows the key findings of our experiments, which compares the performance of \method{} to the three other baseline approaches. 
The $x$-axis indicates the time each method takes to complete a single task, while the $y$-axis indicates the corresponding user rating.
The results demonstrate that our \method{} achieves the highest user rating and the shortest execution time compared to the other approaches. 
While GROP proves to be as efficient as our approach, it receives a significantly lower rating score. 
By contrast, \random{} and \LLM{} both receive lower user ratings than our \method{}.
They also display poor efficiency.
This is because they lack the navigation capabilities to efficiently navigate through complex environments.
For instance, when their navigation goals are located within an obstacle area, they struggle to adjust their trajectory, leading to longer task completion times.

\begin{figure*}
\begin{center}
    \includegraphics[width=0.97\textwidth]{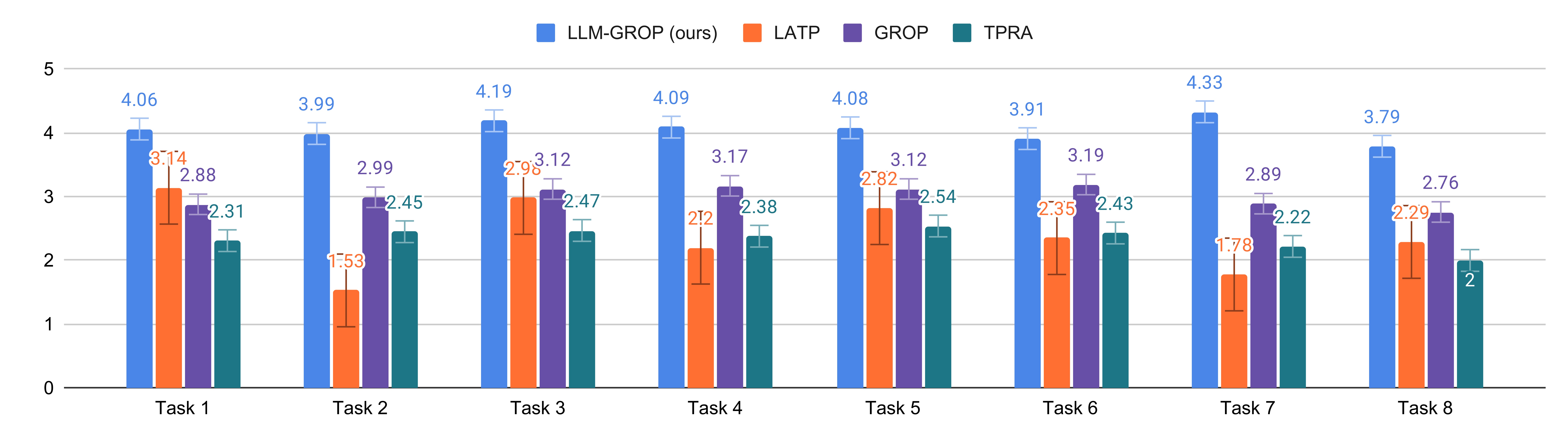}
    \caption{User ratings of individual object rearrangement tasks, with the $x$-axis representing the task and the $y$-axis representing the user rating score. It can be observed that \method{} consistently performs the best compared to baselines. Tasks 1-5 involve three objects, tasks 6 and 7 involve four objects, and task 8 involves five objects. The numerical value displayed on each bar indicates the mean rating for the corresponding task.
    }\label{fig:individual}
\end{center}
\end{figure*}

Figure~\ref{fig:individual} presents the individual comparison results of each method for individual tasks. 
The $x$-axis corresponds to Task \#ID in Table~\ref{table:task}, while the $y$-axis represents the average user rating for each method.
Our \method{} demonstrates superior performance over the baselines for each task. 
Specifically, tasks 1 to 5 receive slightly higher scores than tasks 6 and 8. 
This is reasonable because the latter two tasks require the robot to manipulate more objects, posing additional challenges for the robot.

\subsection{Implementations with Different LLMs}

An important component of \method{} is an off-the-shelf LLM. 
To evaluate the sensitivity of \method{} to the choice of LLMs, we conducted experiments by realizing \method{}s with four different LLMs. 

\noindent
\textbf{Experimental Setup:} 
We evaluated four versions of our system using different LLMs: GPT-3, GPT-4, Gemini, and Claude.
These models were selected for their diverse architectures and capabilities, enabling us to assess their respective influences.
Another group of 10 participants with engineering backgrounds, comprising four females and six males, was recruited to evaluate the method.
We applied the same rating criteria used in the performance assessment.

\noindent
\textbf{Results:} 
The results of this experiment, presented in Table~\ref{table:ablation result}, revealed that GPT-4 outperformed other models in four out of eight tasks, demonstrating robust performance across varying task complexities.
Notably, GPT-4 achieved the highest average user rating scores for Tasks 1, 3, 6, and 7, which involved object counts ranging from three to four.
This consistency indicates that GPT-4 is well-suited for managing both simpler and moderately complex object rearrangement tasks with reliable performance.

However, the results also highlight the strengths of other models in specific tasks, underscoring their varied capabilities depending on task characteristics.
For instance, Gemini achieved the highest scores in Tasks 2 and 4, suggesting a particular aptitude for tasks with unique relational or spatial requirements.
Similarly, Claude outperformed the other models in Task 5, indicating potential advantages in handling intricate or nuanced task relationships within a three-object setup.

Interestingly, in Task 8—the most complex task involving five objects—GPT-3 received the highest score.
This variation in performance demonstrates that no single model consistently excels across all task types.
As such, these findings emphasize the importance of selecting the most appropriate model based on the specific relational and structural demands of each task.

\begin{table*}[t]
\centering
\caption{Average of user ratings score for individual object rearrangement tasks. Tasks 1-5 involve three objects, tasks 6 and 7 involve four objects, and task 8 involves five objects. Results of four different LLMs (including their model names) are reported in four different columns. 
}

\label{table:ablation result}
\scriptsize
\begin{tabular}{c|c|c|c|c}
\toprule
	  & \textbf{GPT-3 (text-davinci-003)} & \textbf{GPT-4 (gpt-4-0613
)} & \textbf{Gemini (gemini-1.5-flash-20240520)} & \textbf{Claude (claude-3.5-sonnet-20240620)}\\ 
    \midrule
     Task 1 & $4.06\pm 0.75$ & $\mathbf{4.98\pm 0.04}$ & $4.87\pm 0.18$ & $4.50\pm 0.00$ \\
    \midrule
     Task 2 & $3.99\pm 0.30$ & $4.48\pm 0.42$ & $\mathbf{4.70\pm 0.16}$ & $3.65\pm 0.85$ \\ 
    \midrule
     Task 3 & $4.19\pm 0.30$ & $\mathbf{4.60\pm 0.40}$ & $2.95\pm 0.05$ & $2.83\pm 0.00$ \\ 
    \midrule
     Task 4 & $4.09\pm 0.22$ & $3.15\pm 1.27$ & $\mathbf{4.83\pm 0.00}$ & $4.65\pm 0.04$ \\ 
    \midrule
     Task 5 & $4.08\pm 0.31$ & $4.67\pm 0.00$ & $2.27\pm 0.07$ & $\mathbf{4.73\pm 0.04}$ \\
    \midrule
     Task 6 & $3.91\pm 0.60$ & $\mathbf{4.92\pm 0.00}$ & $2.36\pm 0.07$ & $4.90\pm 0.04$ \\
    \midrule
     Task 7 & $4.33\pm 0.23$ & $\mathbf{4.98\pm 0.04}$ & $4.65\pm 0.14$ & $3.72\pm 1.34$ \\
    \midrule
     Task 8 & $\mathbf{3.79\pm 0.31}$ & $2.22\pm 0.60$ & $1.83\pm 0.34$ & $1.00\pm 0.00$ \\
\bottomrule
\end{tabular}
\end{table*}

\section{Conclusion and Discussion}
To summarize, we propose \method{}, which demonstrates how we can extract semantic information from LLMs and use it as a way to make common sense, semantically valid decisions about object placements as a part of a task and motion planner - letting us execute multi-step tasks in complex environments in response to natural-language commands. 
Further, \method{} leverages computer vision methods to visually ground task and motion planning (TAMP) solutions for mobile manipulation (MoMa) tasks, and enable the robot to select standing positions to simultaneously facilitate both navigation and manipulation actions. 
\method{} was evaluated through comparisons with existing TAMP methods, both in simulation and on a real robot, and results demonstrated its superiority in MoMa tasks. 

\noindent
\textbf{End-to-end robot control:}
One important contribution of this research is the introduction of a novel computer vision approach to help a mobile manipulator select base positions. 
Such positions are not too close to the table, so the robot's base does not have physical contact with the table or other obstacles; at the same time, the standing positions should not be too far away from the table, so interaction behaviors with the target object(s) on the table are secured with high feasibility. 
Our discussion assumes fixed motion controllers for the mobile base and the manipulator, whereas in practice, one can customize those controllers for different tasks and different robot platforms. 
For one example, a biped humanoid robot might be able to squeeze through the crowded to reach a table, which is impossible for a chubby wheeled robot. 
For another, grasping a heavy beer pitcher would require a robot to stand close to the table compared with lighter objects such as forks and knives. 
Incorporating such control-level capabilities into the framework for \method{} practitioners would be one step closer to producing globally optimal TAMP solutions. 

\noindent
\textbf{Open worlds:} The real world is generally open and there can be innumerable situations that are unforeseen in the development of robot hardware and software. 
In MoMa tasks, a robot needs to move around obstacles and rearrange objects into a goal configurations. 
The target objects and obstacles can be novel, and the service requests for specifying goal configurations can be novel too.
LLM-GROP assumes that complete 3D models exist for every object in the scene. In practice, we pick real-world objects whose geometry and appearance closely match our simulation assets. Although one could employ Real2Sim methods, such as 3D reconstruction or image-to-3D generative models, to build accurate object models, we leave those extensions for future work.

TAMP methods frequently assume known objects and known robot capabilities. 
As a result, finding TAMP solutions at both task and motion levels in open-world scenarios is still an open question. 
Foundation models are equipped with rich common sense information and can be potentially useful for addressing those open-world questions. 
While this work demonstrated that LLMs are useful for computing semantically specified object configurations, future work can further look into the capability of foundation models (LLMs and multimodal models) to better embrace the openness of the real world.

\noindent
\textbf{Natural language input:}
Recent advancements in foundation models including LLMs have made it possible for service robots to accurately interpret open-vocabulary, potentially ambiguous natural language inputs. 
In doing so, the robots need to process language and non-language inputs at the same time, and reliably build associations between them, e.g., associating the tokens of ``banana'' to one physical object on a specific dining table. 
Natural language often involves under-specified and incomplete task specifications, as it is unrealistic to expect people to specify every detail of a goal configuration. 
The robot has at least the two options of asking clarification questions to seek additional details of the service request and using common sense to provide the service that makes the best sense. 
Using common sense to make decisions, e.g., people like coffee in the morning, can lead to efficient user experience, while the downside is the hallucinations of user preferences or even risks to people and environments. 
The robot needs to make such decisions based on its AI alignment, utility functions and value systems. 

We used commercially available LLMs in this research including GPTs, Claude and Gemini, so the expenses go beyond time and energy, while further extending to monetary costs of querying the LLMs. 
Such costs are not discussed in the experiments. 
One future work direction can be in the minimization of costs in the usage of LLMs. 

\noindent
\textbf{Ego-centric vision:}
To compute the standing positions for optimally supporting the navigation and manipulation actions, the robot needs an estimation of the world configurations. 
In this paper, we use top-down view images for perceiving the object locations and the world configurations. 
There is rich literature in pose estimation of objects in 3D worlds. 
One future work direction is to incorporate ego-centric vision for estimating world configurations, and active perception would even further enhance the robot's state estimation capability. 
Recent vision-language models (such as LLaVA and GPT-4o) can potentially be used as navigation goal selectors, and their performance can potentially be improved by multimodal prompting techniques like Set-of-Mark (SoM)~\citep{yang2023set}. 

\noindent
\textbf{Simultaneous manipulation and navigation:}
In this research, manipulation and navigation behaviors are temporally interleaved, and they are connected through the selection of standing positions. 
In general case, mobile manipulators can choose to do both at the same time leveraging whole-body control methods. 
Simultaneous manipulation and navigation is an open problem to robotics researchers, where the main challenge is the curse of dimensionality in control space. 
While interleaving manipulation and navigation is sufficient for MoMa robots to complete most object rearrangement tasks, there are scenarios where doing both is a must. 
For instance, when a robot and a human move an object together, the robot needs to adjust its grasping strategy while co-navigating with the human. 
Another example is to push a plate from one end of a long banquet table to the other. 
Those behaviors are still uncommon for current service robots in household environments, but generally can be useful for MoMa tasks. 

\noindent
\textbf{Movable objects:}
Many objects in the real worlds are movable and their movability depends on the robot's MoMa skills. 
Among those movable objects, humans are special because their locations are changed through interaction actions such as language and gestures, instead of contact-rich behaviors such as grasping and pushing. 
This research assumes objects are stationary except for those target objects that are involved in the rearrangement tasks. 
For generalization, future work can incorporate interaction behaviors, such as saying ``excuse me'', for encouraging people movements. 
In addition, there is existing research on TAMP methods for domains with movable objects, where a robot can do second-order reasoning to move obstacle objects to facilitate the manipulation of target objects.

\section*{Acknowledgment}
A portion of this work has taken place at the Autonomous Intelligent Robotics (AIR) Group, SUNY Binghamton. AIR research is supported in part by the NSF (NRI-1925044, IIS-2428998), DEEP Robotics, Ford Motor Company, OPPO, Guiding Eyes, and SUNY RF. 
A portion of this work has taken place in the Learning Agents Research
Group (LARG) at UT Austin.  LARG research is supported in part by NSF
(FAIN-2019844, NRT-2125858), ONR (N00014-24-1-2550), ARO
(W911NF-17-2-0181, W911NF-23-2-0004, W911NF-25-1-0065), DARPA
(Cooperative Agreement HR00112520004 on Ad Hoc Teamwork) Lockheed
Martin, and UT Austin's Good Systems grand challenge.  Peter Stone
serves as the Chief Scientist of Sony AI and receives financial
compensation for that role.  The terms of this arrangement have been
reviewed and approved by the University of Texas at Austin in accordance
with its policy on objectivity in research.

\bibliographystyle{SageH}
\bibliography{ref}

\end{document}